\documentclass[a4paper]{article}

\usepackage{INTERSPEECH2022}
\usepackage{color}
\renewcommand{\paragraph}[1]{\noindent\textbf{#1}\quad}
\usepackage{url}
\usepackage{hyperref}
\newcommand{\code}{\texttt}
\usepackage{xspace}
\usepackage{pifont}
\newcommand{\cmark}{\ding{51}}

\newcommand{\method}{PTL\xspace}
\usepackage{multirow}
\usepackage{booktabs}

\title{Leveraging Pseudo-labeled Data to Improve Direct Speech-to-Speech Translation}
\name{Qianqian Dong$^{1,*}$, Fengpeng Yue$^{2,*, \dag}$, Tom Ko$^{1}$,  Mingxuan Wang$^1$, Qibing Bai$^{2, \dag}$, Yu Zhang$^{2,3}$ \thanks{*~Equal contribution.} \thanks{\dag~Work done during internship at Bytedance. } }
\address{
  $^1$ByteDance AI Lab\\
  $^2$Department of Computer Science and Engineering, \\
Southern University of Science and Technology, Shenzhen, China \\
  $^3$Peng Cheng Laboratory, Shenzhen, China}
\email{\{dongqianqian,tom.ko,wangmingxuan.89\}@bytedance.com,
        \{11930381,12032871\}@mail.sustech.edu.cn,
        yu.zhang.ust@gmail.com}

\begin{document}

\maketitle
\begin{abstract}
\label{sec:abstract}

Direct Speech-to-speech translation (S2ST) has drawn more and more attention recently. 
The task is very challenging due to data scarcity and complex speech-to-speech mapping. 
In this paper, we report our recent achievements in S2ST.
Firstly, we build a S2ST Transformer baseline which outperforms the original Translatotron. 
Secondly, we utilize the external data by pseudo-labeling and obtain a new state-of-the-art result on the Fisher English-to-Spanish test set. 
Indeed, we exploit the pseudo data with a combination of popular techniques which are not trivial when applied to S2ST.
Moreover, we evaluate our approach on both syntactically similar (Spanish-English) and distant (English-Chinese) language pairs.
Our implementation is available at \url{https://github.com/fengpeng-yue/speech-to-speech-translation}.

\noindent\textbf{Index Terms}: speech translation, speech-to-speech translation, pseudo-labeling

\end{abstract}

\section{Introduction}
\label{sec:introduction}

The speech-to-speech translation (S2ST) task translates speech from the source language into speech in the target language.
The conventional S2ST is implemented by a cascaded system \cite{lavie1997janus}, which includes three components: automatic speech recognition (ASR), text-to-text machine translation (MT), and text-to-speech synthesis (TTS).
Like cascaded speech-to-text translation (ST), the two main shortcomings of cascaded S2ST are time delay and error accumulation.
The end-to-end (E2E) approach, which jointly optimizes all components with a single model, effectively alleviates these problems.
With the success of end-to-end ST, the community starts to move on to direct S2ST, a more difficult task.

Translatotron \cite{jia19_interspeech} is one of the pioneering works in direct S2ST, and it is conducted by adopting multi-task learning \cite{zy21}.
It verifies that the end-to-end method can obtain reasonable translation quality and generate intelligible speech.
After that, \cite{jia2021translatotron} proposes Translatotron2 to improve the robustness of the predicted speech and retain the source speaker’s voice in the translated speech better.
For distant language pairs, \cite{kano2021transformer} explores the multi-step training with Transcoder.
The pre-trained MT and TTS encoders are used as the teacher model to facilitate direct S2ST in learning complex linguistic and modal transitions. 

On the other hand, \cite{lee2021direct} aims to build a direct S2ST for unwritten languages. Instead of predicting continuous spectrograms, this work predicts discrete units learned from self-supervised representations of the target speech.
Multi-task learning can be conducted either with text data or not.
Furthermore, a textless S2ST system is proposed by \cite{lee2021textless} and can be trained without any text data.
At the same time, it can generate multi-speaker target speech by training on real-world S2ST data.

One of the challenges to train the end-to-end S2ST system is data scarcity, as collecting speech translation data is expensive.
To tackle this challenge, \cite{duquenne2021multimodal} proposes an automatic data mining method to perform speech-to-speech mining.
However, no work has proven that it is feasible to train the S2ST on these automatically mined data.
Meanwhile, the pseudo-labeled data effectively improve the performance of E2E ST  \cite{jia2019leveraging, mccarthy2020skinaugment} when real-world data is limited. 
However, currently there are few investigations on direct S2ST.
This motivates us to investigate leveraging a large amount of pseudo-labeled data to enhance the performance for S2ST when only limited paired data can be obtained.
As more and more large-scale ASR resources are open-sourced, in this paper, we utilize well-trained MT and TTS model to convert ASR data to pseudo-labeled S2ST data.
To evaluate different methods for using pseudo-labeled data, we conduct experiments on similar language pairs (i.e., Spanish-English) and distant language pairs (i.e., English-Chinese).
Our contributions are three-fold:
\begin{itemize}
    \item We build a strong Transformer-based Translatotron baseline that outperforms the original Translatotron. 
    We report the thorough evaluation of the effects of hyperparameter tuning.
    
    \item We examine the effectiveness of using pseudo-labeled data with pre-training and different fine-tuning strategies.
    By utilizing pseudo-labeled data and the prompt-tuning technique~\cite{liu2021pre}, our best model achieves new state-of-the-art results on the Fisher dataset \cite{post2013improved}.
    
    \item Except for Spanish-English, we evaluate our approach on English-Chinese, which is a syntactically distant language pair, to facilitate the S2ST research in different language communities.

\end{itemize}

The rest of the paper is organized as follows. Section 2 describes our implementation of Transformer-based Translatotron. In Section 3, we introduce our pseudo-labeling approach. Section 4 describes the experiments and presents the analysis.
Section 5 concludes our work.

\section{Transformer-based Translatotron}
\label{sec:e2e_s2s}
\begin{figure}[h]
\begin{center}
    \includegraphics[width=6.7cm]{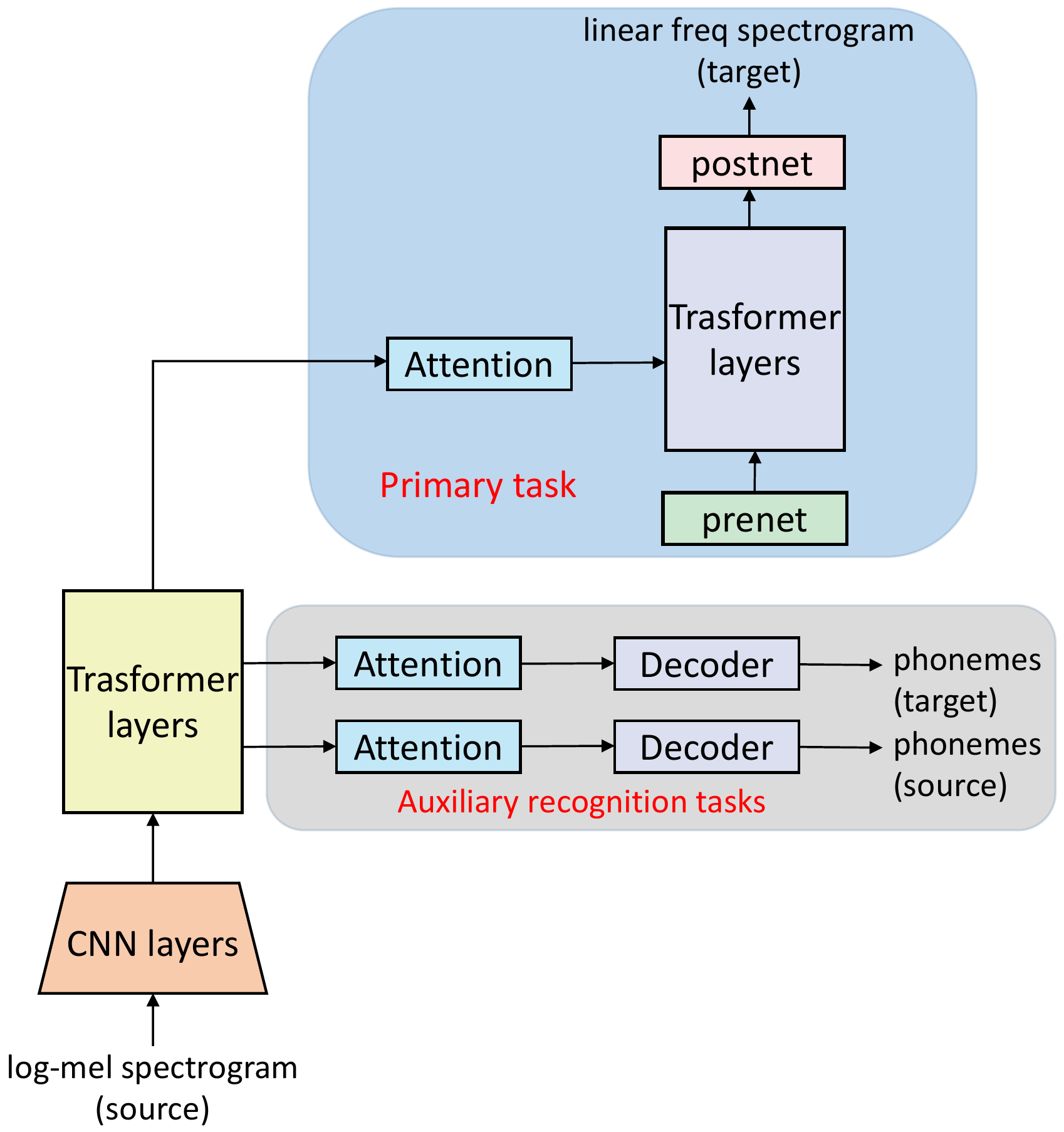}
\end{center}
\caption{Transformer-based Translatotron translates speech from the source language (bottom left) into the speech of the target language (top right).
The auxiliary tasks help learn the speech-to-speech translation. 
Compared with the original Translatotron, we focus on generating target speech on a single speaker  without using the speaker encoder. }
\label{Figure:model}
\end{figure}

As shown in Figure \ref{Figure:model}, our model structure follows the Translatotron \cite{jia19_interspeech}. 
We adopt the Transformer \cite{vaswani2017attention} instead of LSTM as the backbone.
We find that Transformer-based  Translatotron can get much better performance than original Translatotron with careful parameter tuning.

\subsection{Transformer Encoder}
Subsampling the speech feature in the encoder is an effective way to help the model pick up attention during training.
In our model, the 80-channel mel-spectrogram input features are subsampled to a quarter of the original size by two convolutional layers and then fed into the Transformer layer. 
The Transformer encoder has 12 Transformer layers with 512-dimension hidden units in each layer and 8 heads in each multi-head attention block.
For the position-wise feed-forward networks, we use 2048 dimensional inner states.
During training, SpecAugment \cite{park19e_interspeech} is applied as the data augmentation strategy. 

\subsection{Transformer Decoder}

Similar to the Transformer TTS \cite{li2019neural}, the decoder that predicts the spectrogram in the target language includes pre-net, Transformer layers, and post-net components.
By following the setting in Translatotron, the dimension of the pre-net bottleneck is set to 32.
Based on the performance on the development set, we set the reduction factor \cite{wang2017tacotron} to 4 for the output feature frames.
The Transformer decoder has 6 Transformer layers with the same hyperparameters as the encoder's Transformer layers.

\subsection{Multi-task Learning}
Following Translatotorn, we also employ multi-task learning training strategy.
In addition to the primary task (i.e., speech-to-speech translation), two additional tasks (i.e., speech recognition and speech-to-text translation) are included in Translatotron. 
The two auxiliary tasks play an important role in Translatotron, and they are conducted by two additional decoders.
The auxiliary decoders take the intermediate hidden representation of the encoder to predict phoneme sequences.
Intuitively, the shallow encoder layers represent the source linguistic content, while the deep layers encode more information about the target linguistic content.
During training, the weighted auxiliary losses are added to the overall training loss.
In Section 4, we demonstrate that the performance of the primary task is sensitive to the hyperparameters of the auxiliary tasks.
Table \ref{tab:parameters} shows the details about the hyperparameter setting of our best practice.

\begin{table}[!ht]
    \centering
     \caption{Model hyperparameters for Fisher and TedEn2Zh datasets.}
    \begin{tabular}{lcc}
    \toprule
    \textbf{\#Param.} &\textbf{Fisher} & \textbf{TedEn2Zh} \\
    \midrule
     Input / output sample rate (Hz) & 8k/24k& 16k/24k \\
     Transformer Encoder   & 12 x 512 & 12 x 512 \\
     Transformer Decoder   & 6 x 512 & 6 x 512\\
     Auxiliary Transformer Decoder & 1 x 64 & 4 x 64\\
     ~~source / target encoder layer   & 6 / 9 & 4 / 9\\
     ~~source / target loss weight     & 0.3 / 0.3 & 0.3 / 0.3 \\
    Learning rate         & 0.006 & 0.0015\\
    Warmup steps        & 4000 & 4000\\
    Dropout        & 0.1 & 0.1 \\
    Batch tokens    & 60000 & 45000\\
    \bottomrule
    \end{tabular}
    \label{tab:parameters}
\end{table}

\section{Pseudo Translation Labeling}
\label{sec:pseudo_label}
As real S2ST data is very limited nowadays, pseudo-labeling is an intuitive approach to alleviate the problem. Most existing work \cite{jia19_interspeech,jia2021translatotron,kano2021transformer} converts ST data to S2ST data with a TTS system and conducts their experiments. However, real ST data is also limited, and the data scarcity problem remains severe. 
In this paper, we extend the pseudo-labeling approach by converting the ASR data into ST data with a MT system and then into S2ST data. 

In our work, we first create a primary dataset $\mathcal{A}$ from a ST corpus. $\mathcal{A}=\{\mathbf{s_{src}}, \mathbf{t_{src}}, \mathbf{t_{tgt}}, \mathbf{s_{tgt}'}\}$ represents \{real source speech, real transcription, real translation, pseudo target speech\}.
Then we create a secondary dataset $\mathcal{B}$ from an ASR corpus. $\mathcal{B}=\{\mathbf{s_{src}}, \mathbf{t_{src}}, \mathbf{t_{tgt}'}, \mathbf{s_{tgt}'}\}$ represents \{real source speech, real transcription, pseudo translation, pseudo target speech\}.
Here, dataset $\mathcal{B}$ is much larger than $\mathcal{A}$ in scale.
We examine the effectiveness of using the secondary dataset with pre-training and different fine-tuning strategies.
As the target translation sequences of dataset $\mathcal{B}$ are generated by MT, we name our approach as pseudo translation labeling (\method).

\subsection{Pre-training and Fine-tuning}

We first use dataset $\mathcal{B}$ to pre-train the encoder so that the model can learn a better representation.
The pre-training is done with an equal weight for two tasks: ASR and ST, which make use of $\{\mathbf{s_{src}}, \mathbf{t_{src}}\}$ and $\{\mathbf{s_{src}}, \mathbf{t_{tgt}'}\}$, respectively.
The two tasks share representations from the same encoder but with different decoders.
The pre-trained encoder and the decoders are reused in the upcoming fine-tuning stage.
Dataset $\mathcal{A}$ is then used to optimize the overall model to carry out the basic fine-tuning.

\subsection{Mixed-tuning}

In order to further improve the model, we conduct fine-tuning with a mixture of dataset $\mathcal{A}$ and $\mathcal{B}$.
As dataset $\mathcal{B}$ is much larger than $\mathcal{A}$, we duplicate the samples in the primary set to balance the overall distribution.
In this work, we assume that the distribution of the target test set is close to the distribution of the primary training set.
Thus, we explicitly apply upsampling to prevent the model from getting biased to the secondary training set.

\subsection{Prompt-tuning}

In order to enhance the ability of the model to learn the difference between various data sources, we adopt the ``pre-train, prompt, and predict"~\cite{liu2021pre} paradigm.
Based on pre-training, we take the category of the datasets (including ``\code{$<$primary$>$}" and ``\code{$<$secondary$>$}") as a prompt, and attach it to the input features of each sample in the form of the predefined embeddings during the prompt-tuning stage. 
With the explicit textual prompt, we can manipulate the model's behavior in the inference stage.

\section{Experiments}
\label{sec:experiments}
\subsection{Datasets}

We conduct experiments on two language pairs (i.e., Spanish to English and English to Chinese).
The former belongs to the same language family, while the latter belongs to a different language family.
We construct S2ST paired data based on the two ST datasets, Fisher Spanish\cite{post2013improved} and TedEn2Zh\cite{liu2019end}, by using the in-house TTS to synthesize the target speech from the translation sequences.
We utilize the in-house MT to convert ASR data that includes a subset of Gigaspeech \cite{GigaSpeech2021} (Giga-Sub) and the Spanish subset of multilingual LibriSpeech \cite{pratap2020mls} (MLS\_Es)
to pseudo-labeled ST data, and then use the same in-house TTS to synthesize the target speech from the pseudo translation sequences.
As described in Section~\ref{sec:pseudo_label}, for the Spanish-to-English language pair, Fisher and MLS\_Es are the primary and secondary datasets, respectively.
TedEn2Zh and Giga-Sub are the the primary and secondary datasets for the English-to-Chinese language pair, respectively.
Table \ref{tab:dataset} shows the statistics of each dataset.

\begin{table}[!ht]
    \centering
    \setlength\tabcolsep{1pt}
    \caption{Statistics of training data for Spanish-to-English (Es $\rightarrow$ En) and English-to-Chinese (En $\rightarrow$ Zh). ``Mixed" includes audiobook, podcast, and YouTube. 
    All duration statistics are based on segmented audios.}
    \begin{tabular}{lcccc}
     \toprule
    \multirow{2.5}*{\textbf{Dataset}}   & \multicolumn{2}{c}{\textbf{Es $\rightarrow$ En}}  &
    \multicolumn{2}{c}{\textbf{En $\rightarrow$ Zh}} \\
    \cmidrule(r){2-3}
    \cmidrule(r){4-5}
      & primary & secondary & primary & secondary \\  
    \midrule
      Data source   & Fisher & MLS\_Es & TedEn2Zh & Giga-Sub \\
    
      Source hour &171 & 918 & 524 & 1566\\
      Target hour &146 & 738 & 467&1478\\
    Utterance &130K &220K &294K & 1.1M\\
    Sampling rate & 8 kHz& 16 kHz & 16 kHz & 16 kHz\\
      Domain & Conversation& Audiobook &Lecture &Mixed\\
    \bottomrule
    \end{tabular}
    \label{tab:dataset}
\end{table}

\subsection{Implementation Details}
\paragraph{Data Preprocessing}
To be consistent with the sampling rate (SR) of Fisher, we downsample MLS\_Es to 8 kHz.
Our acoustic features are 80-channel mel-spectrogram extracted from source speech as input and 80-channel mel-spectrogram extracted from target speech as output.
We use phonemes from the source and target languages as modeling units for the two auxiliary tasks. 
We filter out the code-switched texts and those texts for which TTS fails to generate speech.

\paragraph{Evaluation}
Following previous work, we use the BLEU score as an objective evaluation metric to measure the translation accuracy and the mean opinion score (MOS) as a subjective evaluation metric to measure the naturalness of pronunciation for predicted target speech. 
We utilize the pre-trained ASR to recognize the predicted target speech and then calculate the BLEU score between the resulting transcripts and ground-truth reference translations.
For English ASR, we use the \code{Wav2vec2}\cite{baevski2020wav2vec} and \code{CTC} model\footnote{\url{https://huggingface.co/facebook/wav2vec2-large-960h-lv60-self}} from \code{Huggingface}\cite{wolf-etal-2020-transformers}, which is pretrained and fine-tuned on Libri-Light\cite{kahn2020libri} and  960 hours of Librispeech\cite{panayotov2015librispeech}  corpus.
For Chinese ASR, we use the attention-based \code{Conformer}\cite{gulati2020conformer} model\footnote{\url{https://github.com/wenet-e2e/wenet/tree/main/examples/aishell}} from \code{Wenet}\cite{yao2021wenet} trained on the AISHELL\cite{bu2017aishell} corpus. 
We report case-insensitive detokenized BLEU scores calculated by \code{sacrebleu}\footnote{\url{https://github.com/mjpost/sacrebleu}} on Fisher dataset.
Meanwhile, we report character-level BLEU scores on the TedEn2Zh dataset.

\paragraph{Cascade S2ST}
Our cascade S2ST is built by cascading E2E ST and TTS.
We train a Transformer-based E2E ST following the model setting in \cite{inaguma2020espnet} with ASR pre-training.
In the cascade S2ST system, we take the predicted translation of ST to generate the target speech using the in-house TTS.

\subsection{Main Results}
In this section, we illustrate the effectiveness of our method from the perspective of objective evaluation.
Translatotron-T represents the Transformer-based Translatotron in all tables.
\subsubsection{Objective Evaluation}

\begin{table}[!ht]
    \centering
    \setlength\tabcolsep{1.5pt}
    \caption{Performance on the dev and test sets of Fisher Spanish-English dataset. 
    }
    \begin{tabular}{lccc}
    \toprule
     \textbf{Method}  & \textbf{dev-BLEU} & \textbf{dev2-BLEU}&
     \textbf{test-BLEU}\\
    \midrule
     Translatotron~\cite{jia2019direct}     &24.8&26.5& 25.6\\
     ~~+~Encoder PT    &30.1&31.5&31.1\\
     Translatotron-2~\cite{jia2021translatotron}    &-&-& 37.0\\
     ~~+~ConcatAug    &-&-& 40.3\\
    UWSpeech~\cite{zhang2021uwspeech}   &-&-&9.4\\
    S2UT~\cite{lee2021direct}   &-&-& 39.9\\
     Translatotron-T     &32.1&32.8&32.0\\
     ~~+~\method    &\textbf{42.4} &\textbf{43.3}&\textbf{43.6}\\
    \midrule
    Cascade S2ST~\cite{jia2019direct}    &39.4&41.2&  41.4\\
    Cascade S2ST~\cite{jia2021translatotron}    &-&-& 43.3\\
    Cascade S2ST   &44.3&45.4&45.1\\
    \midrule
    Ground truth~\cite{jia2019direct} &82.8&83.8&  85.3\\
    Ground truth~\cite{jia2021translatotron}&-&-&  88.6\\
    Ground truth &88.1&88.6& 89.8\\
    \bottomrule
    \end{tabular}
    \label{tab:main_fisher}
\end{table}

\paragraph{Performance on Fisher dataset}
We present our results in Table \ref{tab:main_fisher}.
It demonstrates that our baseline results surpass the original Translatotron results on Fisher. 
Moreover, we adopt the \method  described in Section 3 on our baseline, which improves the BLEU score significantly.
Overall, our method can achieve an improvement of nearly 3 BLEU scores compared with the previous best end-to-end model, Translatotron-2 with ``ConcatAug'' data augmentation. 

In Table~\ref{tab:main_fisher_2}, we report the performance on auxiliary tasks under different methods. 
The results show that the auxiliary tasks and the primary task have a positive correlation in performance. 
This indicates that using external data to improve the performance of auxiliary tasks will also benefit the S2ST, which is consistent with the motivation of pre-training auxiliary tasks.

\begin{table}[!ht]
    \centering
    \setlength\tabcolsep{3pt}
    \caption{Performance of auxiliary tasks on the test set of Fisher Spanish-English dataset. ``S-PER" means phoneme error rate (PER) of auxiliary ASR task on test set. ``Tp-BLEU" means phone-based BLEU of auxiliary ST task on test set.}
    \begin{tabular}{lccc}
    \toprule
     \textbf{Method}  & \textbf{S-PER($\downarrow$)} & \textbf{Tp-BLEU} & 
    \textbf{ test-BLEU}\\
    \midrule
     Translatotron-T    &16.10 &55.68 &32.0 \\
     ~~+~\method    & \textbf{13.95}&\textbf{ 61.96}&\textbf{43.6} \\
    \bottomrule
    \end{tabular}
    \label{tab:main_fisher_2}
\end{table}

\paragraph{Performance on TedEn2Zh dataset}
English-Chinese translation is a more difficult task because the two languages are much different in grammar and syntax.
As shown in Table~\ref{tab:main_teden2zh}, we report the results of Transform-based Translatotron on the Ted2Zh dataset.
The BLEU score of the baseline is merely 11.2 because the predicted results contain a large amount of unintelligent speech.
The \method can bring about 7 points of BLEU improvement, which significantly enhances translation accuracy.
Our best model even surpasses the cascade system.

\begin{table}[!ht]
    \centering
    \setlength\tabcolsep{2pt}
    \caption{Performance on the dev and test set of TedEn2Zh dataset. ``S-PER" means PER of auxiliary ASR task on test set. ``Tp-BLEU" means phoneme-based BLEU of auxiliary ST task on test set. 
    }
    \begin{tabular}{lcccc}
    \toprule
     \textbf{Method}  & \textbf{S-PER($\downarrow$)} & \textbf{Tp-BLEU} & \textbf{dev-BLEU}&
     \textbf{test-BLEU}\\
    \midrule
     Translatotron-T    & 13.06&44.57 &7.0 & 11.2\\
     ~~+~\method    &\textbf{11.26} & \textbf{50.73}& \textbf{12.2}& \textbf{20.8}\\
    \midrule
    Cascade S2ST    & -& -& 11.8& 19.7\\
    \midrule
    Ground truth &- &- & 82.9 & 94.9\\
    \bottomrule
    \end{tabular}
    \label{tab:main_teden2zh}
\end{table}

\paragraph{Parameters for Auxiliary Task} 
We explore the influence of the hyperparameters of auxiliary tasks on the performance of S2ST on the Fisher Spanish dataset, and the results are shown in Table~\ref{tab:bleu_parameter}. 
We find that the performance of the model with the size of the smaller auxiliary decoder is better.
We conjecture that it will force the encoder shared by the auxiliary task and the primary task to learn more helpful information, which is more conducive to the training of the primary task.

\begin{table}[!ht]
    \centering
    \caption{Effect of the parameters for the auxiliary tasks (ASR, ST) on the test set of Fisher dataset. ``\#Pos" represents the source and target encoder layer. ``\#Num" and ``\#Dim" mean the number and dimension of the Transformer layers of two auxiliary decoders , respectively. }
    \begin{tabular}{ccccc}
    \toprule
      \textbf{Exp.}  & \textbf{\#Pos} & \textbf{\#Num} & \textbf{\#Dim} & \textbf{test-BLEU} \\
    \midrule
     I   & (4,9) & (4,4) & (128,128) & 24.2\\
     II &(4,9)&(3,3)&(128,128)& 25.7\\
     III &(6,9)&(3,3)&(128,128)& 28.4\\
     IV &(6,9)&(1,1)&(64,64)& \textbf{32.0}\\
    \bottomrule
    \end{tabular}
    \label{tab:bleu_parameter}
\end{table}

\paragraph{Effect of different methods }
As shown in Table~\ref{tab:bleu_pseudo}, we compare different methods of the pseudo translation labeling approach. 
When we apply the pre-training to the auxiliary tasks by pseudo-labeled data (Method-I), the BLUE scores significantly improve compared with the baseline.
Further, based on pre-training, mixed-tuning (Method-II) improves 2.8 BELU on Fisher and 5.5 BELU on TedEn2Zh.
As shown in Table~\ref{tab:dataset}, there is an obvious mismatch between the primary data and the secondary data in the two language pairs.
The prompt-tuning (Method-III) helps the model distinguish different data sources, and further gains can be obtained on both language pairs.

\begin{table}[!ht]
    \centering
    \setlength\tabcolsep{2pt}
    \caption{Comparison on the effectiveness of the three methods. The BLEU scores are reported on the test sets of Fisher and TedEn2Zh datasets. ``Pre-training" is conducted on the dataset $\mathcal{B}$ for all three methods.}
    \begin{tabular}{lcccc}
    \toprule
     \textbf{Model}   & \textbf{Pre-training}& \textbf{Fine-tuning}& \textbf{Fisher} & \textbf{TedEn2Zh} \\
    \midrule
    Translatotron-T &-&-&32.0 & 11.2 \\
     Method-I &\cmark & \cmark&39.2 & 14.8\\
     Method-II &\cmark&Mixed-tuning&43.0 & 20.3\\
     Method-III &\cmark&Prompt-tuning&\textbf{43.6} & \textbf{20.8}\\
    \bottomrule
    \end{tabular}
    \label{tab:bleu_pseudo}
\end{table}

\subsubsection{Subjective Evaluation}
In Table \ref{tab:mose_main}, we use the Hifi-GAN \cite{kong2020hifi} vocoder to synthesize audios from predicted spectrograms and conduct the MOS test to evaluate the naturalness of audios.
The gain from our \method approach on MOS is consistent with BLEU.
In particular, our method significantly enhances the intelligibility of audios on the TedEn2Zh dataset (English-Chinese).

\begin{table}[!ht]
    \centering
    \caption{Naturalness MOS Evaluation on test sets of the two datasets. The ground truth for both datasets are synthetic target speech from the in-house TTS.}
    \begin{tabular}{llcc}
    \toprule
     \textbf{Method} & \textbf{Fisher} & \textbf{TedEn2Zh}\\
    \midrule
     Translatotron-T  & 3.40$\pm 0.16$ & 2.59$\pm 0.18$ \\
     ~~+~\method         & \textbf{3.59$\pm 0.17$} & \textbf{3.20$\pm 0.18$} \\
     
    \midrule
     Ground truth & 3.87$\pm 0.19$ & 3.95$\pm 0.18$ \\
     \bottomrule
    \end{tabular}
    \label{tab:mose_main}
\end{table}

\section{Conclusions}
\label{sec:conclusion}
In this paper, we first build a strong Transformer-based Translatotron baseline for direct S2ST, which obviously outperforms the original Translatotron. 
As the S2ST performance is sensitive to the hyperparameters of the auxiliary decoder, we have made careful tuning to get the best performance.
Furthermore, to tackle data scarcity, we examine the effectiveness of employing pseudo-labeling with pre-training and various fine-tuning strategies.
Our system can achieve new state-of-the-art performance on Fisher (Spanish-English) dataset.
Finally, we report the performance of TedEn2Zh (English-Chinese) dataset to facilitate the direct S2ST research on more language pairs.

\bibliographystyle{IEEEtran}
\bibliography{main}

\end{document}